\def\year{2022}\relax
\documentclass[letterpaper]{article} 
\usepackage{aaai22}  
\usepackage{times}  
\usepackage{helvet}  
\usepackage{courier}  
\usepackage[hyphens]{url}  
\usepackage{graphicx} 
\usepackage{natbib}  
\usepackage{caption} 
\DeclareCaptionStyle{ruled}{labelfont=normalfont,labelsep=colon,strut=off} 
\usepackage{algorithm}
\usepackage{makecell}
\usepackage{algorithmic}
\usepackage{array}

\usepackage{epsfig}
\usepackage{amsmath}
\usepackage{amssymb}
\usepackage{multirow}
\usepackage{caption}
\usepackage{cite}
\usepackage{color,xcolor}
\usepackage[colorlinks,
            linkcolor=black,       
            anchorcolor=black,  
            citecolor=black,        
            ]{hyperref}
\usepackage[switch]{lineno}

\newcommand{\eg}{\mbox{e.g.,\ }}

\newcommand{\ie}{\mbox{i.e.,\ }}

\usepackage{newfloat}
\usepackage{listings}
\floatstyle{ruled}
\newfloat{listing}{tb}{lst}{}
\floatname{listing}{Listing}
\title{Text Gestalt: Stroke-Aware Scene Text Image Super-Resolution}

\author{
    Jingye Chen$^{1}$, Haiyang Yu$^{1}$, Jianqi Ma$^{2}$, Bin Li\protect\thanks{Corresponding author}$^{1}$, Xiangyang Xue$^{* 1}$
    \\
}
\affiliations{
 $^{1}$Shanghai Key Laboratory of Intelligent Information Processing\\
 School of Computer Science, Fudan University\\
 $^{2}$The Hong Kong Polytechnic University \\ 
 \{jingyechen19, hyyu20, libin, xyxue\}@fudan.edu.cn, jianqi.ma@connect.polyu.hk
}

\usepackage{bibentry}

\begin{document}
\maketitle

\begin{abstract}
     In the last decade, the blossom of deep learning has witnessed the rapid development of scene text recognition. However, the recognition of low-resolution scene text images remains a challenge. Even though some super-resolution methods have been proposed to tackle this problem, they usually treat text images as general images while ignoring the fact that
    the visual quality of strokes (the atomic unit of text) plays an essential role for text recognition. According to Gestalt Psychology, humans are capable of composing parts of details into the most similar objects guided by prior knowledge. Likewise, when humans observe a low-resolution text image, they will inherently use partial stroke-level details to recover the appearance of holistic characters. Inspired by Gestalt Psychology, we put forward a Stroke-Aware Scene Text Image Super-Resolution method containing a Stroke-Focused Module (SFM) to concentrate on stroke-level internal structures of characters in text images. Specifically, we  attempt to design rules for decomposing English characters and digits at stroke-level, then pre-train a text recognizer to provide stroke-level attention maps as positional clues with the purpose of controlling the consistency between the generated super-resolution image and high-resolution ground truth. The extensive experimental results validate that the proposed method can indeed generate more distinguishable images on TextZoom and manually constructed Chinese character dataset Degraded-IC13. Furthermore, since the proposed SFM is only used to provide stroke-level guidance when training, it will not bring any time overhead during the test phase. Code is available at 
    \href{https://github.com/FudanVI/FudanOCR/tree/main/text-gestalt}{https://github.com/FudanVI/FudanOCR/text-gestalt}.
   
\end{abstract}

\section{Introduction}
In recent years, scene text recognition has achieved tremendous progress owing to the rapid development of deep learning. It has been widely used in many real-world applications such as auto-driving \cite{zhang2020street}, ID card recognition \cite{satyawan2019citizen}, signature identification \cite{ren2020st}, etc. Although the recently proposed recognizers become stronger as reported, we observe that low-resolution (LR) text images still pose great challenges for them. In this context, a super-resolution module is required as a pre-processor to recover the missing details of LR images. 

\begin{figure}[t]
    \centering
    \includegraphics[width=0.46\textwidth]{ 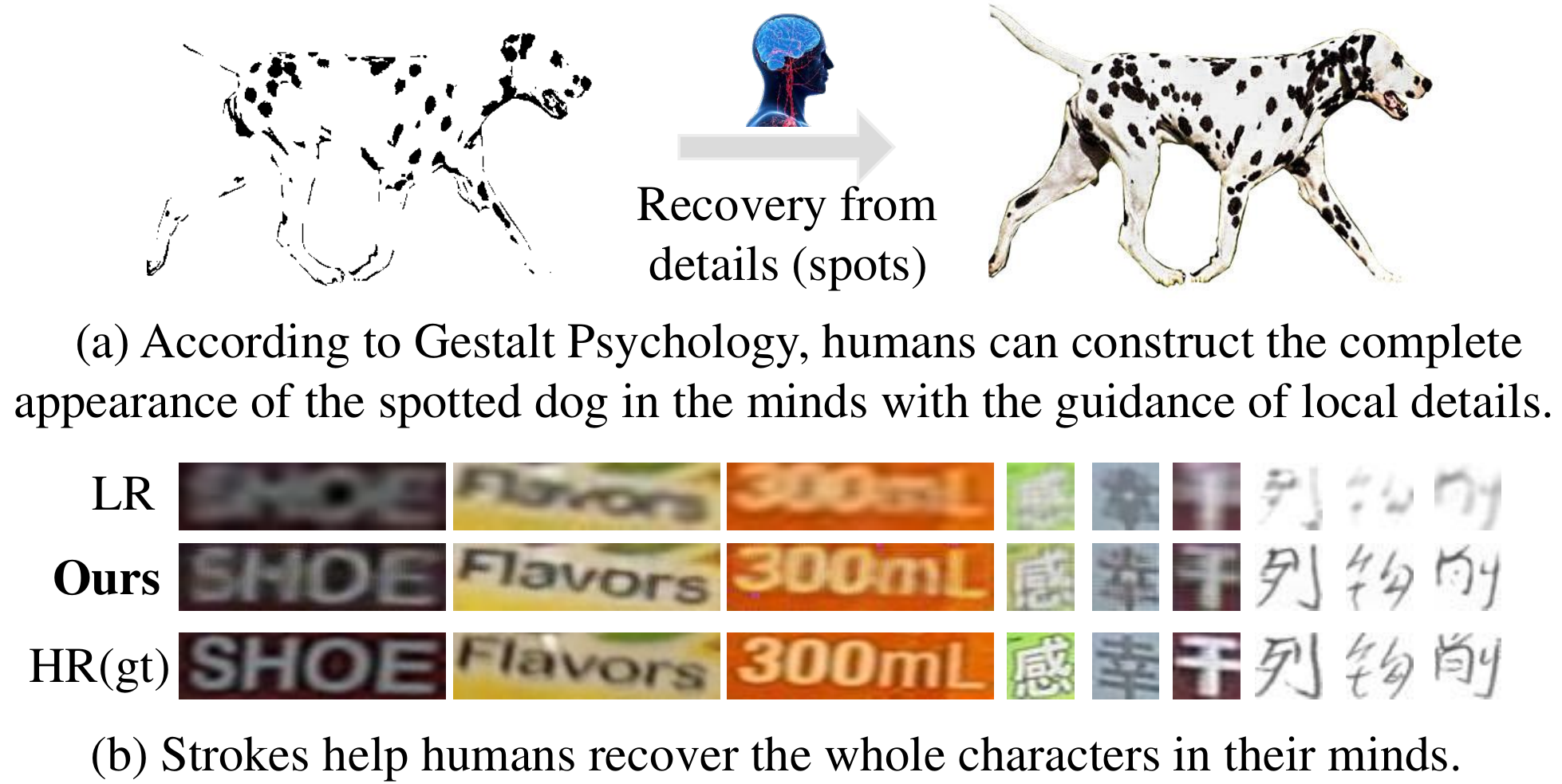}
    \caption{For incomplete or blurred images, detailed information (\eg spots or strokes) play a significant role during recovery. Our method can generate recognizable English and Chinese text images with the guidance of stroke details.}
    \label{fig:dog}
\end{figure}

The previous super-resolution methods usually try to learn degradation patterns through HR-LR pairs with global loss functions (\eg L1 or L2 loss) to recover every pixel in text images \cite{xu2017learning,pandey2018binary}. These methods, however, usually view text images as general images regardless of text-specific properties. Recently, a few methods attempt to take several text-specific properties into account, which achieve better performance in terms of both image quality and recognition accuracy. For example, PlugNet \cite{yanplugnet} employs a multi-task framework with the purpose of jointly optimizing super-resolution and text recognition tasks in one model. In \cite{wang2020scene}, the authors introduced a Text Super-Resolution Network (TSRN) via appending two recurrent layers in the backbone to capture sequential information of text images. The recently proposed Scene Text Telescope (STT) \cite{chen2021scene} introduces text priors into the model by proposing a position-aware module and a content-aware module. The concurrent work TPGSR \cite{ma2021text} incorporates text-specific semantic features to each block in the backbone and exerts an iterative way to enhance text images. 
Through observations, the text priors used in these works usually regard \textbf{\textit{character}} as the smallest unit of text lines, whereas ignoring the significance of more detailed internal structures. In this paper, we take a step further to answer the critical question: \emph{Can text priors at a more fine-grained level (\eg \textbf{stroke}) benefit the super-resolution procedure?}

According to \textbf{Gestalt Psychology} \cite{kohler1967gestalt}, humans can compose parts of details into the most similar objects guided by prior knowledge. As is shown in Figure \ref{fig:dog}(a), humans can inherently recover the whole appearance of the spotted dog with the guidance of local details such as spots. Likewise, for blurred text images, strokes that act as local details play an indispensable role in the recovery process. As is shown in Figure \ref{fig:dog}(b), even though the character ``m'' in ``300ml'' looks blurred, we can easily recover it when discovering the three parallel vertical strokes.

Inspired by Gestalt Psychology, we propose a stroke-aware Scene Text Image Super-Resolution method that utilizes a Stroke-Focused Module (SFM) to take advantage of fine-grained stroke-level attention maps generated by an auxiliary recognizer as guidance for recovery. Different from most existing recognizers \cite{shi2016end,shi2018aster,luo2019moran} that predict at character-level, we design a recognizer working at the stroke level, thus is capable of generating more fine-grained attention maps. To validate the effectiveness of our method, we employ some recognizers and image quality metrics to evaluate the generated SR images. The experimental results show that our method can indeed achieve state-of-the-art performance on the TextZoom and designed Chinese character dataset Degraded-IC13 in terms of recognition accuracy. Moreover, since the proposed SFM is only used when training, it will not bring any time overhead during testing. Our contributions are listed as follows:
\begin{itemize}
    \item We attempt to design rules for recognizing English letters and digits at the stroke level to provide more fine-grained attention-level guidance. 
    \item Inspired by Gestalt Psychology, we propose a Stroke-Focused Module (SFM) to concentrate more on stroke regions with the guidance of stroke-level attention maps.
    \item 
    Compared to the previous methods, our method can generate more distinguishable text images on the TextZoom and Degraded-IC13 in terms of recognition accuracy without bringing any time overhead during testing.
\end{itemize}


\section{Related Work}
\begin{figure*}[t]
    \centering
    \includegraphics[width=1.0\textwidth]{ 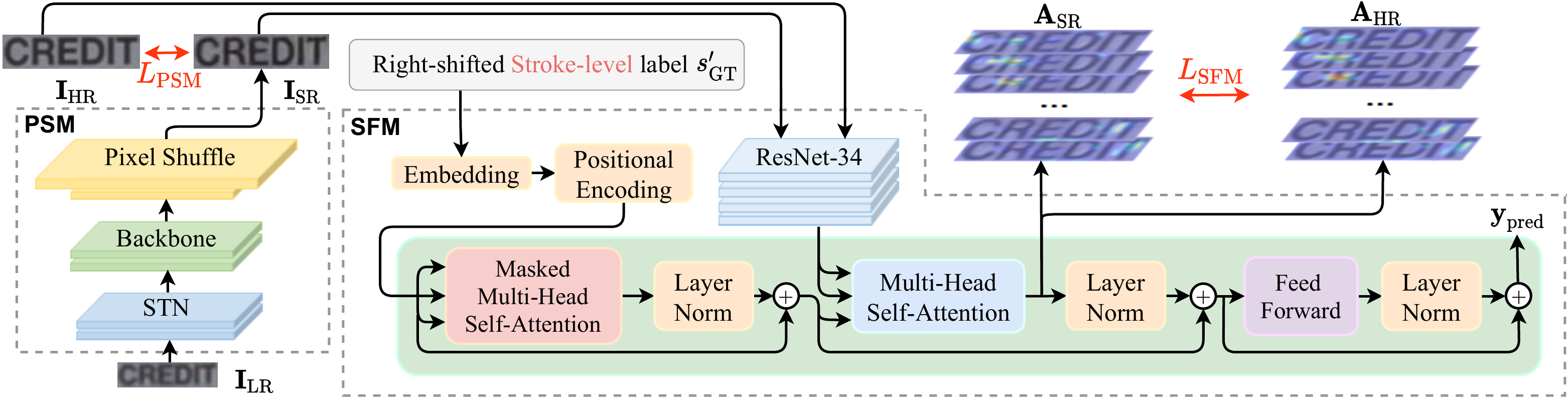}
    \caption{The overall architecture of our method. It contains two modules, including a Pixel-wise Supervision Module (PSM) to recover the color and contour of text images and a Stroke-Focused Module (SFM) to highlight the details of stroke regions.}
    \label{fig:architecture}
\end{figure*}



\subsection{Single Image Super-Resolution}
Single image super-resolution aims to generate an SR image based on its LR counterpart while recovering several missing details. In the deep learning era, the first CNN-based method named SRCNN \cite{dong2014learning} establishes an end-to-end approach to learning the mapping from LR to HR images using a shallow network, while achieving better performance compared with previous traditional methods. EDSR \cite{lim2017enhanced} proposes a deep model by using multiple residual blocks for better representation and removing several unnecessary batch normalization layers in the residual blocks. MSRN \cite{li2018multi} introduces filters of different sizes in two branches while extracting multi-scale features.

\subsection{Text Image Super-Resolution}
Traditional methods usually utilize classical machine learning algorithms to upsample LR images. In \cite{capel2000super}, a Maximum \textit{a posteriori} approach combined with a Huber prior was applied to TISR. In \cite{dalley2004single}, a Bayesian framework was proposed to upsample binary text images. However, the design of traditional features was time-consuming and the low-capacity features were subpar to tackle such task \cite{chen2021text}. Recently, PlugNet \cite{yanplugnet} designs a multi-task framework by optimizing recognition and super-resolution branches in one model. To capture sequential information of text images, in \cite{wang2020scene}, the authors proposed a TSRN containing two BLSTMs. STT \cite{chen2021scene} contains two text-focused modules including a position-aware module and a content-aware module providing text priors. TPGSR \cite{ma2021text} combines text priors in the encoder and employs an iterative manner to enhance low-resolution images. However, these methods usually view characters as the smallest units without considering the more fine-grained details like strokes.  

\subsection{Scene Text Recognition}
Traditional methods usually adopt a bottom-up approach to recognize text images \cite{ wang2010word,wang2011end,neumann2012real}. Specifically, they first detect and classify separated characters and then compose them into text lines with the guidance of language models or lexicons. In the deep learning era, CRNN \cite{shi2016end} combines CNN and RNN as the encoder and employs a CTC-based decoder \cite{graves2006connectionist} to maximize the probability of paths that can reach the ground truth. ASTER \cite{shi2018aster} introduces a Spatial Transformer Network (STN) \cite{jaderberg2015spatial} to rectify irregular text images in an unsupervised manner for better recognition. SEED \cite{qiao2020seed} tries to capture global semantic features of text images with the guidance of a pre-trained fastText model. Although the semantics-based methods are capable of tackling those images with local missing details such as occlusion, they still have difficulty in recognizing low-resolution images with global missing details. Therefore, a preprocessor is required for recovering the details of low-resolution images.

\section{Methodology} 
In this section, we introduce two modules and the way to decompose characters. At last, we introduce the overall loss function. The overall architecture is shown in Figure \ref{fig:architecture}.

\subsection{Pixel-wise Supervision Module}
The existing super-resolution backbones usually follow this design: \textbf{(1)} Employ a series of stacked CNN layers to build up a backbone for extracting features, whose height and width are the same as the original images while containing more channels; \textbf{(2)} Utilize a pixel shuffle module containing multiple CNN layers to reshape the generated maps. Consequently, a super-resolution image is generated with a larger size. The widely used backbones contain SRCNN \cite{dong2014learning}, SRResNet \cite{ledig2017photo}, TSRN \cite{wang2020scene}, TBSRN \cite{chen2021scene}, etc. Please note that there may exist a misalignment problem between LR-HR pairs \cite{wang2020scene}. For example, in the TextZoom dataset, since the pairs are manually cropped and matched by humans, there are several pixel-level offsets that pose difficulties for the super-resolution methods. Hence, we follow \cite{wang2020scene} to append a STN \cite{jaderberg2015spatial} before the backbone to alleviate this problem. Finally, the PSM module is supervised by an L2 loss. We denote the HR images as $\mathbf{I}_{\text{HR}}$ and the generated SR images as $\mathbf{I}_{\text{SR}}$. The loss is calculated as follows: 

\begin{equation}
L_{\text{PSM}} = || \mathbf{I}_{\text{SR}} - \mathbf{I}_{\text{HR}}||_{2}^{2}
\label{eq:psm_loss}
\end{equation}

\begin{figure*}[t]
    \centering
    \includegraphics[width=1.00\textwidth]{ 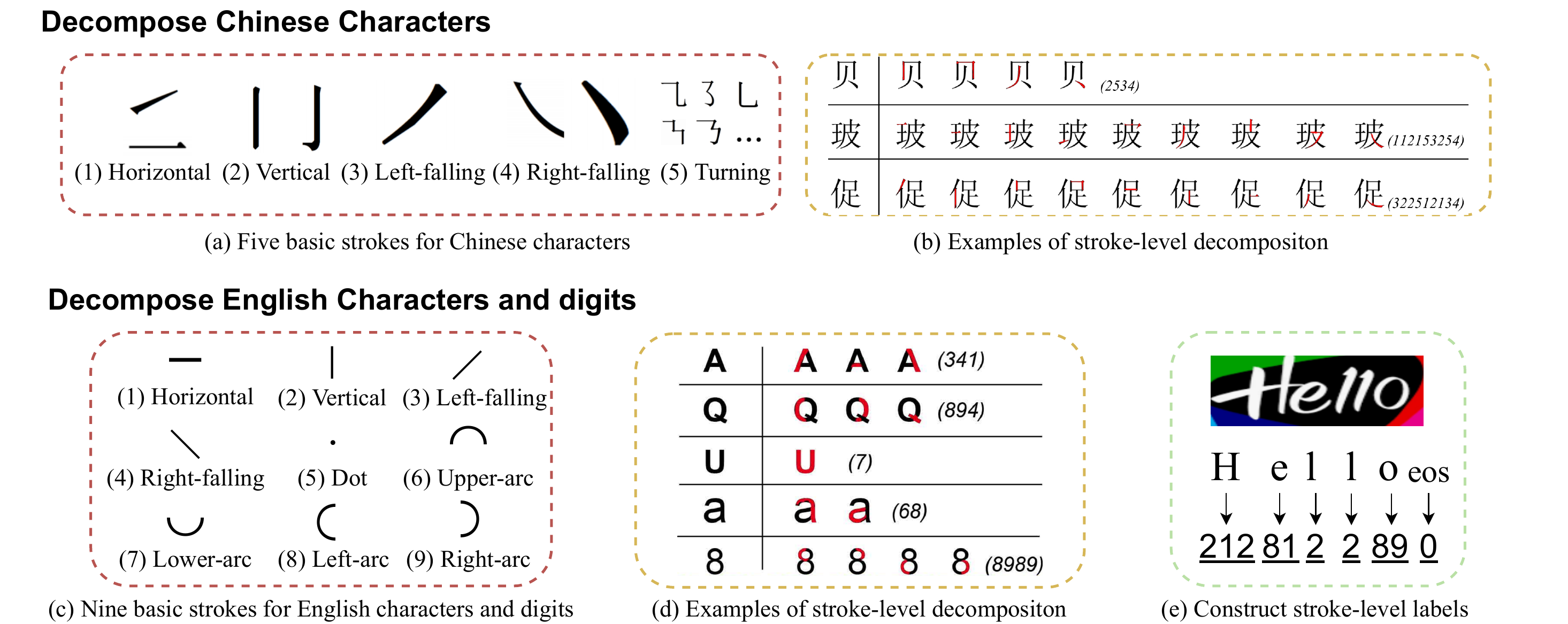}
    \caption{Decomposition of Chinese characters, English characters, and digits. See more examples in Supplementary Material. }
    \label{fig:decompose}
\end{figure*}

\subsection{Stroke-Level Decomposition}
Strokes are the atomic units of characters whatever the language it is. In this section, we try to decompose each character into a stroke sequence and construct stroke-level text labels for English characters, digits, and Chinese characters.

\paragraph{Decompose Chinese characters.}
According to Unicode Han Database, there are five basic strokes of Chinese characters including \textit{horizontal}, \textit{vertical}, \textit{left-falling}, \textit{right-falling}, and \textit{turning}. Each character has a unique stroke sequence and some examples are shown in Figure \ref{fig:decompose}(b). 

\paragraph{Decompose English letters and digits.}
Derived from the approaches to decomposing Chinese characters, we attempt to create stroke encoding for English characters and digits: \textbf{(1)} Break down the characters and digits in more simplified structures, \ie nine basic strokes (see Figure \ref{fig:decompose}(c)). We reduce the total category number for the recognition models to generate better-learned and fine-grained supervision. \textbf{(2)} Represent each character as a sequence of these basic strokes (see Figure \ref{fig:decompose}(d)) \textbf{(3)} Concatenate the stroke sequences of each character and pad a stop symbol ``eos'' in the end (see Figure \ref{fig:decompose}(e)). Please note that we use the category `0' to represent the stop symbol. In this way, we can better make some similar characters distinguishable, \eg `1' and `7' may look similar in some written cases. However, with our stroke encoding, we can denote the character `1' with stroke encoding ``Vertical'' and `7' with stroke encoding ``Horizontal + Vertical'', which can tell the SR model a more fine-grained knowledge for reconstruction. 


\subsection{Stroke-Focused Module}
Strokes perform a significant role in the recognition process. When we see a low-resolution text image, we usually try to capture stroke-level details to infer the appearance of the whole characters in our brain according to Gestalt Psychology \cite{kohler1967gestalt}. Inspired by this, we try to design a module that can provide stroke-level guidance for the super-resolution model. We observe that the existing recognizers \cite{shi2016end,cheng2017focusing,shi2018aster,qiao2020seed} usually regard characters as the smallest units, \ie each character corresponds to a unique class in the alphabet. In this context, recognizers can only attend to coarse-grained character regions at each time step. To exploit more fine-grained attention maps, we pre-train a Transformer-based recognizer on two synthetic datasets, including Synth90k \cite{jaderberg2016reading} and SynthText \cite{gupta2016synthetic} with stroke-level labels following \cite{chen2021zero}. More specifically, given the character-level labels $c_{\text{GT}} = \{c_{1}, c_{2}, ..., c_{t}\}$, we decompose each character and concatenate them to construct the stroke-level labels $s_{\text{GT}} = \{s_{1}, s_{2}, ..., s_{t^{\prime}}\}$, where $t$ and $t^{\prime}$ denote the maximum length of labels at two different levels ($t \leq t^{\prime}$). During pre-training, following \cite{vaswani2017attention,shi2018aster}, we use the force teaching strategy to accelerate the training procedure by employing right-shifted stroke-level label $s_{\text{GT}^{\prime}} = \{s_{<start>}, s_{1}, s_{2}, ..., s_{t^{\prime}-1}\}$ as input, where $s_{<start>}$ denotes the start symbol. We follow the basic design of \cite{chen2021scene} and more details of the encoder and decoder are shown in Supplementary Material. When reaching convergence, we discard the sequence prediction $y_{pred}$ supervised with cross-entropy loss during training, and only leverage the sequence of stroke-level attention maps generated from the Multi-Head Self-Attention Module as stroke-level positional clues. Please note that the parameters in this model are \textbf{frozen} after pre-training. Specifically, we denote the attention maps of HR images as $\mathbf{A}_{\text{HR}} = \{\textbf{A}_{\text{HR}}^{1}, \textbf{A}_{\text{HR}}^{2}, ..., \textbf{A}_{\text{HR}}^{t^{\prime}} \}$ and SR images as $\mathbf{A}_{\text{SR}} = \{\textbf{A}_{\text{SR}}^{1}, \textbf{A}_{\text{SR}}^{2}, ..., \textbf{A}_{\text{SR}}^{t^{\prime}} \}$, then employ an L1 loss to constrain these two maps as follows:

\begin{equation}
L_{\text{SFM}} = || \mathbf{A}_{\text{SR}} - \mathbf{A}_{\text{HR}}||_{1}
\label{eq:sfm_loss}
\end{equation}

\subsection{Overall Loss Function}  
Finally, we construct the overall loss function as follows:
\begin{equation}
L = L_{\text{PSM}} + \lambda_{\text{SFM}} L_{\text{SFM}}
\label{eq:overall_loss}
\end{equation}
where $\lambda_{\text{SFM}}$ balances the weight of these two loss functions. 

\section{Experiments}
In this section, we first introduce the datasets, some evaluation metrics, and implementation details. Then we discuss the choices of parameters. At last, we demonstrate the experimental results.


\paragraph{Datasets.}
The datasets used in this paper are as follows:

\textbf{TextZoom} \cite{wang2020scene} The images in TextZoom originate from RealSR \cite{cai2019toward} and SR-RAW \cite{zhang2019zoom}. These datasets involve LR-HR pairs which are taken by digital cameras in real scenes. Specifically, TextZoom contains $17,367$ LR-HR pairs for training and $4,373$ pairs for testing. In terms of different focal lengths of digital cameras, the test set is divided into three subsets, including $1,619$ LR-HR pairs for the easy subset, $1,411$ LR-HR pairs for the medium subset, and $1,343$ LR-HR pairs for the hard subset. LR images are resized to $16$ $\times$ 64 and HR images are sized to $32$ $\times$ $128$, respectively. Different from handcraft degradation, the LR images in TextZoom suffer from more complicated real-scene degradation, which is more challenging for a certain model to perform the SR text image recovery.

\textbf{IC15} \cite{karatzas2015icdar} contains $1,811$ images originated from natural scenes. It is a challenging benchmark with $352$ images with resolution lower than $16$ $\times$ $64$.

\textbf{Degraded-IC13} is constructed based on IC13-HCCR \cite{yin2013icdar}, which contains 224,419  offline handwritten images covering 3,755 commonly-used Level-1 Chinese characters. Details of the construction are shown in the subsection of Experimental on Degraded-IC13.

\paragraph{Evaluation metrics.} We remove all the punctuations and convert uppercase letters to lowercase letters for calculating recognition accuracy, which follows the setting of \cite{wang2020scene} for a fair comparison. In addition, we use Peak Signal-to-Noise Ratio (PSNR) and Structural Similarity Index Measure (SSIM) to evaluate the quality of SR images.

\paragraph{Implementation details.} Our model is implemented in PyTorch. All experiments are conducted on one NVIDIA GTX 1080Ti GPU with 11GB memory. The model is trained using Adam \cite{kingma2014adam} optimizer with learning rate set to $10^{-4}$. The batch size is set to 16. After pre-trained on Synth90k \cite{jaderberg2016reading} and SynthText \cite{gupta2016synthetic}, the parameters of the Transformer-based recognizers are \textbf{frozen}. SFM is a pluggable module that is only used when training, \ie only the PSM is used to upsample LR images in the test phase.

\subsection{Choices of Parameters}\label{section:choice}

The experiments in this section are all conducted on the TextZoom dataset and we employ CRNN for validation. Specifically, we utilize TSRN as the backbone.

\paragraph{Choices of $\lambda_{\text{SFM}}$.}
$\lambda_{\text{SFM}}$ performs an important role to balance the weight of two loss terms. The higher its value, the more our model focuses on the stroke-level details. We explore the value of $\lambda_{\text{SFM}}$ ranging from \{0, 0.1, 1, 10, 50, 100\} and the experimental results are shown in Table \ref{tab:sensitivity of sfm}. When $\lambda_{\text{SFM}}$ is set to 50, the recognition accuracy reaches the best and it boosts the average accuracy by 7.5\% compared with the baseline ($\lambda_{\text{SFM}}$=0). When it values at a lower level such as 0.1, SFM does not bring much guidance for the module. 
So we set $\lambda_{\text{SFM}}$ to 50 in the following experiments.

\paragraph{Choices of L1 loss and L2 loss.}
Empirically, L1 loss and L2 loss are interchangeable in super-resolution tasks. To further validate the impact of these two losses on the generated images, we conduct experiments on four combinations of them (see Table \ref{tab:combination}). The experimental results show that the performance reaches the best when $L_{\text{PSM}}$ uses L2 loss and $L_{\text{SFM}}$ uses L1 loss. We notice that $L_{\text{SFM}}$ is usually a relatively small value with the order of magnitudes near $10^{-3}$. Hence, it will produce a much smaller gradient when using L2 loss, which is inefficient to supervise the SR learning.

\begin{table}[t]
\caption{Experiments on the choices of $\lambda_{\text{SFM}}$.}
\centering
\scalebox{0.83}{
\begin{tabular}{p{2cm}<{\centering}||p{1.2cm}<{\centering}|p{1.2cm}<{\centering}|p{1.2cm}<{\centering}|p{1.5cm}<{\centering}}
\hline
$\lambda_{\text{SFM}}$ & Easy  & Medium &  Hard & Average\\
\hline
\hline
0 & 52.5\% & 38.2\% & 31.4\% & 41.4\% \\
\hline
0.1 & 56.5\% & 40.9\% & 32.9\% & 44.2\% \\
\hline
1 & 58.9\% & 43.7\% & 34.9\% & 46.6\% \\
\hline
10 & 58.9\% & 46.1\% & 34.4\% & 47.2\% \\
\hline
50 & \textbf{61.2\%} & \textbf{47.6\%} & \textbf{35.5\%} & \textbf{48.9\%} \\
\hline
100 & 60.6\% & 47.7\% & 34.3\% & 48.4\% \\
\hline
\end{tabular}
}
\label{tab:sensitivity of sfm}
\end{table}

\begin{table}[t]
\caption{Experiments on four combinations of two losses.}
\centering
\scalebox{0.9}{
\begin{tabular}{c|c||p{1.2cm}<{\centering}|p{1.2cm}<{\centering}|p{1.2cm}<{\centering}|p{1.2cm}<{\centering}}
\hline 
$L_{\text{PSM}}$ & $L_{\text{SFM}}$ & Easy  & Medium &  Hard & Average\\
\hline
\hline
L1 & L1 & 59.5\% & 46.9\% & 33.9\% & 47.6\% \\
\hline
L1 & L2 & 56.0\% & 42.8\% & 33.3\% & 44.8\% \\
\hline
L2 & L1 & \textbf{61.2\%} & \textbf{47.6\%} & \textbf{35.5\%} & \textbf{48.9\%} \\
\hline
L2 & L2 & 58.9\% & 45.6\% & 34.2\% & 47.0\% \\
\hline
\end{tabular}
}
\label{tab:combination}
\end{table}

\begin{table*}[htb]
\caption{The experimental results on TextZoom (The results of NRTR, SEED, and AutoSTR are in Supplementary Material). The module can generate more recognizable text images with the guidance of SFM. The \textit{\underline{underlined}} numbers indicate the best average accuracy using the specific backbone and recognizer for evaluation. The \textbf{bold} numbers denote the best accuracy.}
\centering
\scalebox{0.81}{
\begin{tabular}{c|c||c c c c|c c c c|c c c c}
\hline 
\multirow{2}*{Backbone} & \multirow{2}*{Focus} &  \multicolumn{4}{c|}{CRNN \cite{shi2016end}} & \multicolumn{4}{c|}{MORAN \cite{luo2019moran}} & \multicolumn{4}{c}{ASTER \cite{shi2018aster}}\\
\cline{3-14}
~ & ~ & Easy & Medium & Hard & Average & Easy & Medium & Hard & Average & Easy & Medium & Hard & Average  \\ 
\hline
\hline 
LR & -  & 36.4\% & 21.1\% & 21.1\% & 26.8\% & 60.6\% & 37.9\% & 30.8\% & 44.1\% & 67.4\% & 42.4\% & 31.2\% & 48.2\%  \\
\hline
HR & - & 76.4\% & 75.1\% & 64.6\% & 72.4\% & 91.2\% & 85.3\% & 74.2\% & 84.1\% & 94.2\% & 87.7\% & 76.2\% & 86.6\%  \\
\hline
\hline
\multirow{3}*{SRCNN}  & None  & 41.1\%  & 22.3\%  & 22.0\%  & 29.2\%  & 63.9\%  & 40.0\%  & 29.4\%  & 45.6\%  & 70.6\%  & 44.0\%  & 31.5\%  & 50.0\%   \\
~ & Char  & 41.7\% & 25.4\% & 23.1\% & 30.7\% & 66.2\% & 44.4\% & 31.3\% & 48.4\% & 70.2\% & 49.4\% & 32.5\% & \underline{51.9\%}  \\
~ & Stroke  &  46.5\% & 30.8\% & 25.2\% & \underline{34.9\%}  & 65.2\% & 46.4\% & 32.2\% & \underline{49.0\%} & 68.8\% & 47.7\% & 33.1\% & 51.0\%  \\
\hline
\multirow{3}*{SRResNet} & None  & 45.2\% & 32.6\% & 25.5\% & 35.1\% & 66.0\% & 47.1\% & 33.4\% & 49.9\% & 69.4\% & 50.5\% & 35.7\% & 53.0\%    \\
~ & Char  & 50.0\% & 36.2\% & 28.4\% & 38.9\% & 70.4\% & 53.9\% & 37.9\% & 55.1\% & 72.6\% & 57.1\% & 38.7\% & 57.2\%  \\
~ & Stroke  & 55.5\% & 42.5\% & 31.2\% & \underline{43.8\%} & 72.9\% & 54.1\% & 36.8\% & \underline{55.7\%} & 74.7\% & 56.2\% & 38.3\% & \underline{57.6\%}  \\
\hline
\multirow{3}*{TBSRN} & None  & 54.2\%  & 40.6\%  & 32.7\%  & 43.2\%  & 71.1\%  & 55.2\%  & 39.5\%  & 56.3\%  & 75.2\%  & 56.8\%  & 40.2\%  & 58.5\%   \\
~ & Char & 59.6\% & 47.1\% & 35.3\% & 48.1\% & 74.1\% & 57.0\% & 40.8\% & \underline{58.4\%} & 75.7\% & 59.9\% & 41.6\% & 60.1\%  \\
~ & Stroke  & \textbf{61.3\%} & 47.2\% & 35.0\% & \underline{48.7\%} & 73.6\% & 57.7\% & 40.3\% & 58.2\%  & 77.4\% & 59.0\% & 41.3\% & \underline{60.4\%}  \\
\hline
\multirow{3}*{TSRN} & None  & 52.5\%  & 38.2\%  & 31.4\%  & 41.4\%  & 70.1\%  & 55.3\%  & 37.9\%  & 55.4\%  & 75.1\%  & 56.3\%  & 40.1\%  & 58.3\%   \\
~ & Char & 54.3\% & 40.4\% & 31.7\% & 42.9\% & 72.3\% & 55.6\% & 39.8\% & 56.9\% & 74.3\% & 59.7\% & 39.6\% & 58.9\%  \\
~ & Stroke  & 61.2\% & \textbf{47.6\%} & \textbf{35.5\%} & \textbf{\underline{48.9\%}} & \textbf{75.8\%} & \textbf{57.8\%} & \textbf{41.4\%} & \textbf{\underline{59.4\%}}  & \textbf{77.9\%} & \textbf{60.2\%} & \textbf{42.4\%} & \textbf{\underline{61.3\%}}  \\
\hline
\end{tabular}}
\label{result on recognizer2}
\label{result on recognizer3}
\label{tab:final result}
\end{table*}

\subsection{Experimental Results}
In this section, we first conduct the experiments on TextZoom, IC15, and Degraded-IC13. When conducting the experiments on TextZoom and IC15, we test with six recognizers of different categories, including CTC-based CRNN \cite{shi2016end}, rectification-based MORAN \cite{luo2019moran} and ASTER \cite{shi2018aster}, Transformer-based NRTR \cite{sheng2019nrtr}, semantics-based SEED \cite{qiao2020seed}, as well as NAS-based AutoSTR \cite{zhang2020efficient}, all of which are available on GitHub in terms of source code and pre-trained weights. When experimenting on Degraded-IC13, we manually train a recognizer on HWDB1.0-1.1 \cite{liu2013online}.

\begin{figure*}[t]
    \centering
    \includegraphics[width=0.97\textwidth]{ 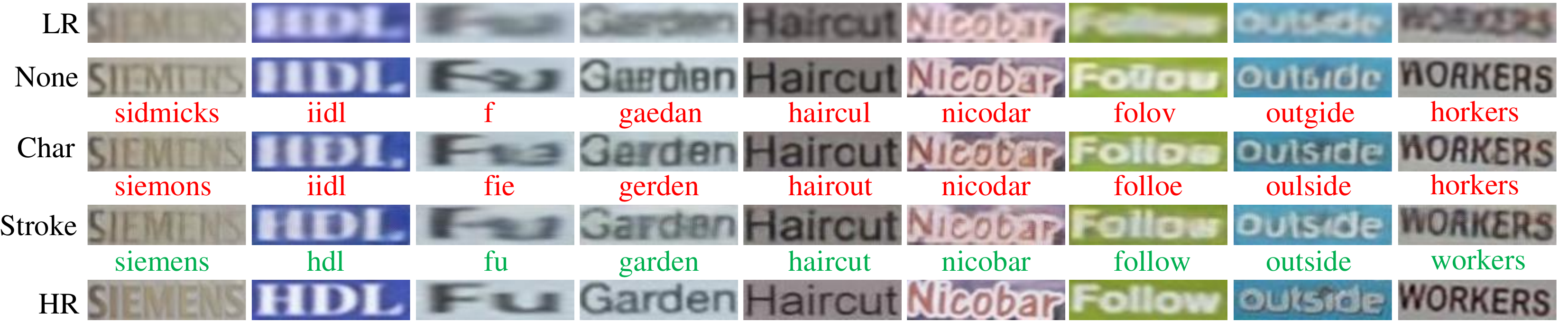}
    \caption{Examples of the generated images. ``None'' means no text priors are taken into account, while ``Char'' and ``Stroke'' denote the model is trained with character-level guidance and stroke-level guidance. We choose TSRN as the backbone.}
    \label{fig:total_example}
\end{figure*}

\paragraph{Experiments on TextZoom.}
The experimental results are shown in Table \ref{tab:final result}. We notice that the proposed SFM can indeed provide positive guidance to boost recognition accuracy. When using TBSRN as the backbone and CRNN for evaluation, the model armed with SFM is capable of boosting the accuracy by 5.5\% and 0.6\% compared with non-focused and character-focused settings. 
%
Since the proposed SFM enhances the HR recovery mainly by concentrating on stroke regions, which may result in less fidelity to the background of the original HR image. Thus, the evaluation metrics like SSIM and PSNR are not stably improved in our cases (see Supplementary Material). To determine the experimental evidence in this situation, we further analyze the trend of $L_{\text{PSM}}$. $L_{\text{PSM}}$ can drop fastly and converge at a relatively lower degree in the absence of SFM. Based on these observations, we come to the following conclusion: \textbf{(1)} PSM usually pays attention to all pixels in given images. Moreover, PSM can perform even better to decrease the super-resolution loss without SFM, thus achieving better scores in terms of two image quality metrics; \textbf{(2)} SFM mainly focuses on stroke-level details, which are separate from background pixels. Intuitively, when we set SFM to large weight ($\lambda_{\text{SFM}}=50$), the model concentrates more on stroke regions while caring less about background pixels, resulting in lower scores on PSNR and SSIM.
However, our aim is to recover recognition-friendly and visual-pleasing text images. The visualization and loss analysis also demonstrate that good PSNR and SSIM scores are not equivalent to well-recovered text images. Moreover, one can clearly see in Figure~\ref{fig:total_example} that, the SR model with SFM supervision demonstrates superior SR text image recovery compared with those without SFM. We also explore the ability of SFM when combined with other character-focused methods, \eg STT \cite{chen2021scene} and TPGSR \cite{ma2021text} (See Table \ref{tab:tpgsr}). We observe that the guidance at character and stroke levels are complementary and the performance can be boosted further when combining text priors at two levels.


\paragraph{Experiments on IC15.}
IC15 \cite{karatzas2015icdar} is one of the widely used English scene text recognition benchmarks. Compared with other datasets such as IC03 \cite{lucas2005icdar} and CUTE80 \cite{risnumawan2014robust}, this dataset contains more incidentally captured images with low resolution, which is a great challenge for the existing recognizers. We manage to validate the ability of the proposed method as a pre-processor. We extract 352 low-resolution images (\ie resolution lower than 16 $\times$ 64) from IC15 as a subset named IC15-352 and test on six recognizers. Please note that we do not use the full dataset since the high-resolution image themselves can be well recognized without super-resolution. We follow three settings, including training TSRN without focus, with character-level focus, and with stroke-level focus. The experimental results are shown in Supplementary Material. The model with stroke-level guidance boosts the accuracy of 3.1\% compared with the model focusing on the character level when evaluated on CRNN. Moreover, when using the guidance of SFM, the accuracy reaches the best in most cases.





\begin{figure}[t]
    \centering
    \includegraphics[width=0.43\textwidth]{ 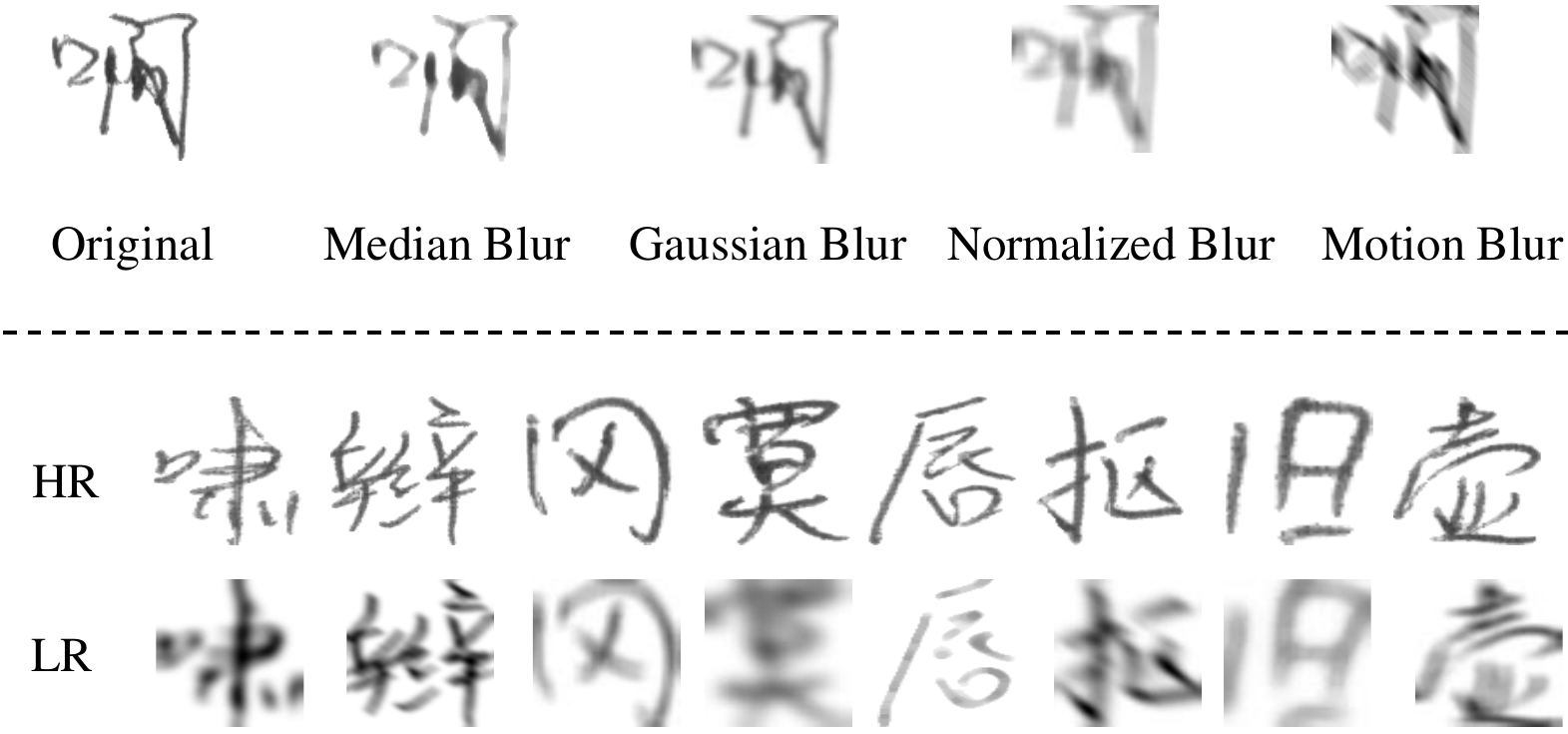}
    \caption{Manually design LR-HR pairs to construct Degraded-IC13. The upper row are four types of blur and the lower row are some examples of LR-HR pairs.}
    \label{fig:chinese}
\end{figure}
\begin{figure}[t]
    \centering
    \includegraphics[width=0.43\textwidth]{ 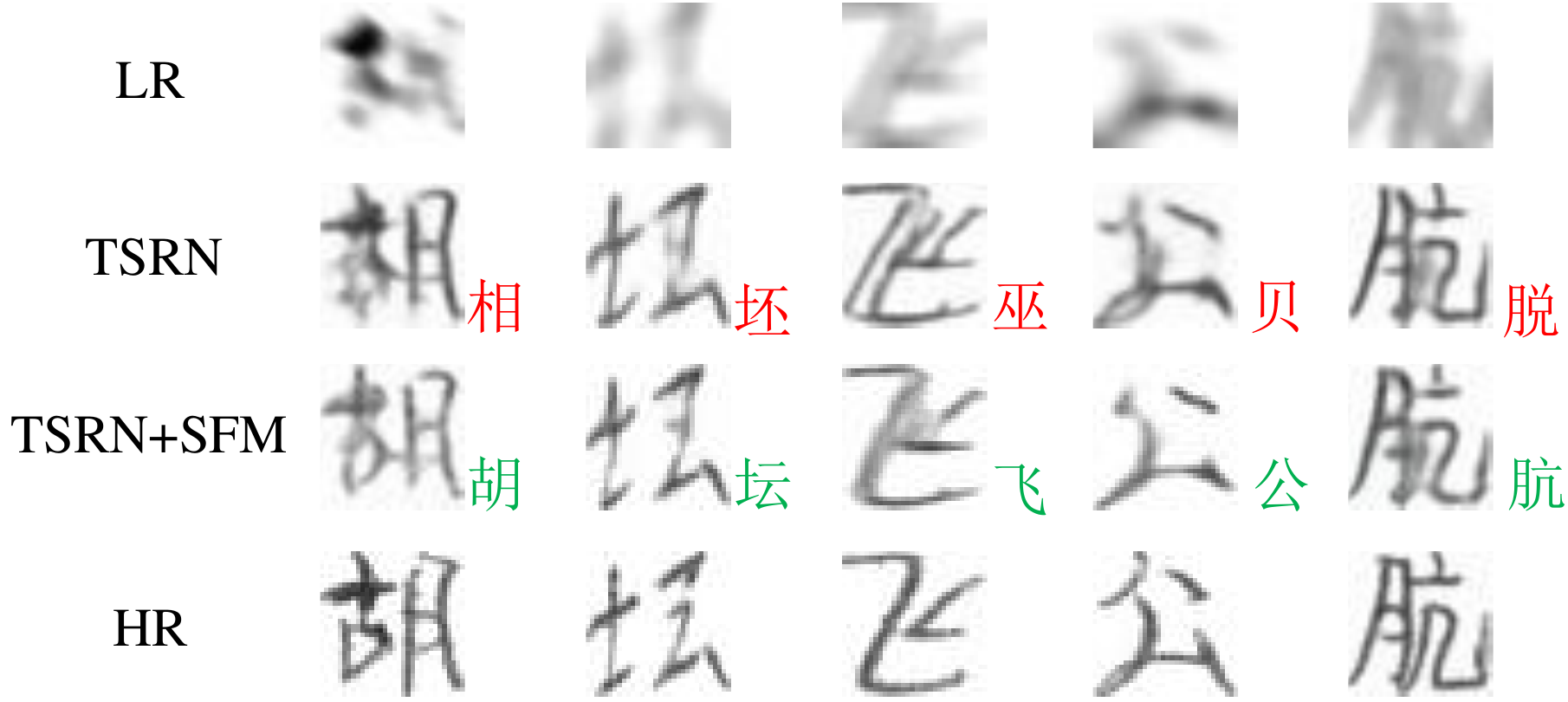}
    \caption{Examples of the generated  Chinese characters.}
    \label{fig:chinese_result}
\end{figure}

\paragraph{Experiments on Degraded-IC13.}\label{sec:ic13}
Compare to English characters, hieroglyph character like Chinese is structured in more complex shape. However, with stroke prior in SFM, we can also equip the SR model capability to recover such complicated characters. To validate the performance of our method on Chinese characters, we construct the Degraded-IC13 dataset in the following ways: \textbf{(1)} We randomly divide IC13-HCCR \cite{liu2013online} into two subsets. Specifically, 179,535 images (80\%) are chosen for training and 44,884 images (20\%) for testing. We first resize them to 64 $\times$ 64; \textbf{(2)} For each image, we randomly select $n$ from {1,2,3,4,5} as the number of blurred operations; \textbf{(3)} We blur the original images for $n$ times. For each time, the blurred type is randomly chosen from four choices, which are demonstrated in Figure \ref{fig:chinese}; \textbf{(4)} We resize the blurred image to 32 $\times$ 32 as LR images using bicubic interpolation. Several examples of the generated HR-LR pairs are demonstrated in Figure \ref{fig:chinese}(b). Following the setting of experiments on English datasets, we pre-train a Chinese recognizer for evaluation and a Transformer-based recognizer that provides stroke-level guidance on the HWDB1.0-1.1 dataset \cite{liu2013online}. The experimental results for the \textit{none-focused}, \textit{character-focused}, and \textit{stroke-focused} are 83.4\%, 84.6\%, 86.1\%, respectively. We notice that the model with SFM can boost the accuracy by almost 3\% when focusing on stroke regions compared with the non-focused setting. Several examples are shown in Figure \ref{fig:chinese_result}. Through the visualizations, we observe that the images generated with the guidance of SFM have relatively clearer strokes, thus achieving better accuracy on the recognizer. 

\begin{table}[t]
\caption{Results of combining TPGSR and STT with SFM. We use TSRN as the backbone and CRNN for evaluation.}
\begin{center}
\scalebox{0.87}{
\begin{tabular}{p{1.2cm}<{\centering}|p{1.2cm}<{\centering}||c|c|c|c}
\hline 
Model & SFM & Easy & Medium & Hard & Average\\
\hline
\hline
\multirow{2}*{TPGSR} & - & 63.1\% & 52.0\% & 38.6\% & 51.8\% \\
\cline{2-6}
~ & \checkmark  & \textbf{64.2\%} & \textbf{53.2\%} & \textbf{38.9\%} & \textbf{52.9\%} \\
\hline
\hline

\multirow{2}*{STT} & - & 61.2\% & 47.6\% & \textbf{35.5\%} & 48.9\% \\
\cline{2-6}
~ & \checkmark  & \textbf{62.3\%} & \textbf{48.1\%} & 35.2\% & \textbf{49.4\%} \\
\hline
\end{tabular}
}
\end{center}
\label{tab:tpgsr}
\end{table}

\begin{table}[t]
\caption{Experimental results on the necessity of the preprocessor. ``TZ'' denotes the training set of TextZoom.}
\centering
\scalebox{0.87}{
\begin{tabular}{l||c|c|c|c}
\hline 
Setting & Easy & Medium & Hard & Average\\
\hline
\hline
(1) Baseline & 45.3\% & 25.4\% & 18.5\% & 30.6\% \\
\hline
(2) Blur-Aug & 53.3\% & 32.7\% & 22.7\% & 37.3\% \\
\hline
(3) Train w/ TZ & 52.1\% & 33.1\% & 22.5\% & 36.9\% \\
\hline
(4) Fine-tune w/ TZ & \textbf{55.4\%} & 35.9\% & 23.6\% & 39.3\% \\
\hline
(5) Preprocessor & 54.8\% & \textbf{40.1\%} & \textbf{28.1\%} & \textbf{41.9\%} \\
\hline
\end{tabular}
}
\label{tab:preprocessor}
\end{table}

\section{Discussions}
\paragraph{Deep insight in pre-trained stroke-level recognizer.}
To provide stroke-level attention maps, the pre-trained recognizer should employ stroke-level text labels unfolded by character-level labels. Before unfolding, the average length of character-level text labels in the training set of TextZoom is 5.0 and the average length of stroke-level text labels reaches 10.9. In fact, attention-based recognizers are easier to suffer from the attention drift problem \cite{cheng2017focusing} when predicting longer sequences. In addition, we notice that after pre-trained on two Synthetic datasets, the recognizer can only achieve 78.0\% recognition accuracy on the training set of TextZoom. Specifically, wrong predictions are usually accompanied by drifted attention maps, which may provide noise for the super-resolution model. To deeply analyze the effect of noise on the SR model, we experiment in three settings: \textbf{(1)} Use attention maps only with \textbf{Correct} predictions. \textbf{(2)} Use \textbf{All} attention maps. \textbf{(3)} Use attention maps only with \textbf{Wrong} predictions. We employ TSRN as the backbone and CRNN for validation. The experimental results are shown in Table \ref{tab:noise}. Interestingly, we observe that the average accuracy of settings (1) and (2) do not show many differences (48.5\% \textit{v.s.} 48.9\%). Based on the result of setting (3), we notice that the performance drops drastically with the wrong guidance. Hence, we come to the conclusion that the stroke-level guidance indeed boosts the performance and the model is robust to resist some disturbances.


\begin{figure}[t]
    \centering
    \includegraphics[width=0.474\textwidth]{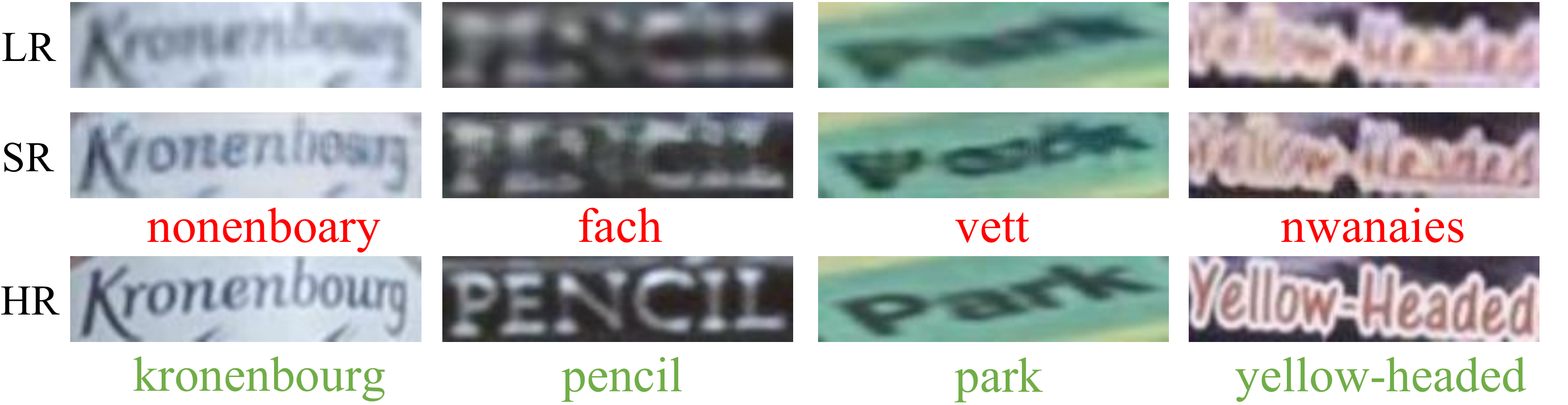}
    \caption{Visualization of some failure cases.}
    \label{fig:failure case}
\end{figure}

\begin{table}[t]
\caption{Experiments on the effect of noise.}
\begin{center}
\scalebox{0.97}{
\begin{tabular}{c||c|c|c|c}
\hline 
\multirow{2}*{Setting} &  \multicolumn{4}{c}{CRNN}  \\
\cline{2-5}
~ & Easy & Medium & Hard & Average  \\ 
\hline
\hline 
(1) Correct & \textbf{61.4\%} & \textbf{47.2\%} & 34.4\% & 48.5\% \\
\hline
(2) All & 61.2\% & \textbf{47.2\%} & \textbf{35.5\%} &  \textbf{48.9\%}  \\
\hline
(3) Wrong & 52.5\% & 35.9\% & 28.6\% & 39.8\% \\
\hline
\end{tabular}}
\end{center}
\label{tab:noise}
\end{table}

\paragraph{Can pre-processor be replaced by training strategies?}
As is mentioned before, the proposed method can indeed boost the recognition performance of existing recognizers on either TextZoom or IC15 datasets. However, here comes a question: \emph{What if the recognizers for evaluation are better trained to adapt to low-quality TextZoom test sets?} To answer this question, we retrain CRNN with Synth90k \cite{jaderberg2016reading} and SynthText \cite{gupta2016synthetic} as the baseline \textit{(Setting 1)}, and utilize some training strategies, including randomly blur synthesize images for data augmentation \textit{(Setting 2)}, combine the HR-LR pairs in the training set of TextZoom with two synthesize datasets for training \textit{(Setting 3)}, fine-tune CRNN with HR-LR pairs in the training set of TextZoom \textit{(Setting 4)}. As is shown in Table \ref{tab:preprocessor}, we observe that the performance reaches the best when the super-resolution model is used as the pre-processor \textit{(Setting 5)}. The reasons may be two folds: \textbf{(1)} The downsampling strategy by algorithms can not simulate the situation in the real scene. \textbf{(2)} By training an additional pre-processor, the model can perceive the degradation process of text images, so it can better generalize to the test dataset.


\begin{table}[t]
\caption{Parameters and FLOPs for two backbones.}
\begin{center}
\scalebox{0.88}{
\begin{tabular}{p{1.4cm}<{\centering}|p{0.8cm}<{\centering}||p{2.5cm}<{\centering}|p{2.5cm}<{\centering}}
\hline
Backbone & SFM & Parameters & FLOPs \\ 
\hline
\hline
\multirow{2}*{SRResNet} & - & 2.5M  & 0.7G  \\ 
~ & \checkmark & 2.5M + 62.0M & 0.7G + 13.6G  \\ 
\hline
\multirow{2}*{TSRN} & - & 2.8M & 0.9G  \\ 
~ & \checkmark & 2.8M + 62.0M  & 0.9G + 13.6G \\ 
\hline
\end{tabular}}
\end{center}
\label{tab:flops}
\end{table}

\paragraph{Can the SR model be extended to other languages?}
We have also conducted experiments on the Korean character dataset PE92 \cite{kim1996handwritten} following the same settings for tackling Chinese characters. The stroke-level decomposition of Korean characters is available in the publicly available code. The accuracy of the Korean recognizer is 90.74\% (none-focused), 90.32\% (character-focused), 92.37\% (stroke-focused), respectively. It further validates the superiority of our method in other languages.

\paragraph{Computational cost.}
In the test phase, we evaluate the time efficiency of our super-resolution method using TSRN as the backbone. We conduct the experment using one NVIDIA GTX 1080TI GPU. To run a batch, the model takes 0.16 seconds without SFM and 0.57 seconds with SFM. The details about parameters and FLOPs are shown in Figure \ref{tab:flops}. In particular, SFM does not bring any time overhead during testing since it is only used in the training phase to provide stroke-level positional clues. 

\paragraph{Failure cases.}
Some failure cases are demonstrated in Figure \ref{fig:failure case}. We observe that our super-resolution method are weak to tackle images with long text since the stroke details are not clear \ie mix with adjacent strokes. Additionally, the oblique text images and images with uneven illumination also bring difficulties to our methods. We will try to mitigate these problems in our future work.

\section{Conclusion}
In this paper, we propose a Stroke-Aware Scene Text Image Super-Resolution method inspired by Gestalt Psychology, highlighting the details on stroke regions. The proposed method can indeed generate more distinguishable super-resolution text images. As is demonstrated in the experimental results, the proposed SFM is capable of achieving state-of-the-art performance on TextZoom and Chinese handwritten datasets without introducing additional time overhead.  

\section{Acknowledgements}
This work was supported in part by the National Natural Science Foundation of China (No.62176060), STCSM project (No.20511100400), Shanghai Municipal Science and Technology Major Project (No.2018SHZDZX01) and ZJLab, Shanghai Research and Innovation Functional Program (No.17DZ2260900), and the Program for Professor of Special Appointment (Eastern Scholar) at Shanghai Institutions of Higher Learning.

\bibliography{main}

\clearpage


\section{Details of Transformer-based Recognizer}
In this section, we introduce the configuration of the proposed stroke-level Transformer-based decoder in detail. Specifically, the decoder receives two kinds of inputs: \textbf{(1)} Right-shifted stroke-level labels input to the Masked Multi-Head Attention module (Masked MHA). \textbf{(2)} Features of images input to the Multi-Head Attention module (MHA). In Masked MHA, the model will attend to the specific regions in feature maps with the guidance of text-level hidden states at each time step. We employ the variant of ResNet-34 \cite{he2016deep} to extract features (see Table \ref{tab:encoder}). For an gray-scale input image with shape of height $H$, width $W$ and channel $C$ ($H \times W \times 1$), the output feature of our encoder is with size of $\frac{H}{4} \times \frac{W}{4} \times 1024$.
We set the hyperparameters of the Transformer decoder following \cite{yang2020holistic}. Specifically, we employ 1 decoder blocks, each of which has 4 heads in attention modules. The dimensionality of positional encoding and embedding are both set to 512. We employ a cross-entropy loss and an Adadelta \cite{zeiler2012adadelta} optimizer to train this model with  learning rate 1.0. 

\begin{table}[h]
\caption{Details of the encoder. \textit{in}, \textit{out}, \textit{kernel}, \textit{stride}, \textit{number} denote input channel, output channel, kernel, stride, and the number of blocks, respectively. The sizes of kernel, stride, padding in CNN are set to 3, 1, 1. A batch normalization layer and a ReLU layer are added after each CNN.}
\begin{center}
\scalebox{0.9}{
\begin{tabular}{cc}
\hline
Layer Name & Configuration \\
\hline
CNN  &  \textit{in}=3, \textit{out}=64 \\
Max-Pooling & \textit{kernel}=2, \textit{stride}=2\\
CNN  &  \textit{in}=64, \textit{out}=128 \\
Max-Pooling & \textit{kernel}=2, \textit{stride}=2\\ 
Building Block & \textit{number}=3, \textit{in}=128, \textit{out}=256 \\
Building Block & \textit{number}=4, \textit{in}=256, \textit{out}=256 \\
Building Block & \textit{number}=6, \textit{in}=256, \textit{out}=512 \\
Building Block & \textit{number}=3, \textit{in}=512, \textit{out}=1024 \\
\hline
\end{tabular}}
\end{center}
\label{tab:encoder}
\end{table}

\section{More Details of Stroke-Level Decomposition}
\begin{figure*}[h]
    \centering
    \includegraphics[width=0.9\textwidth]{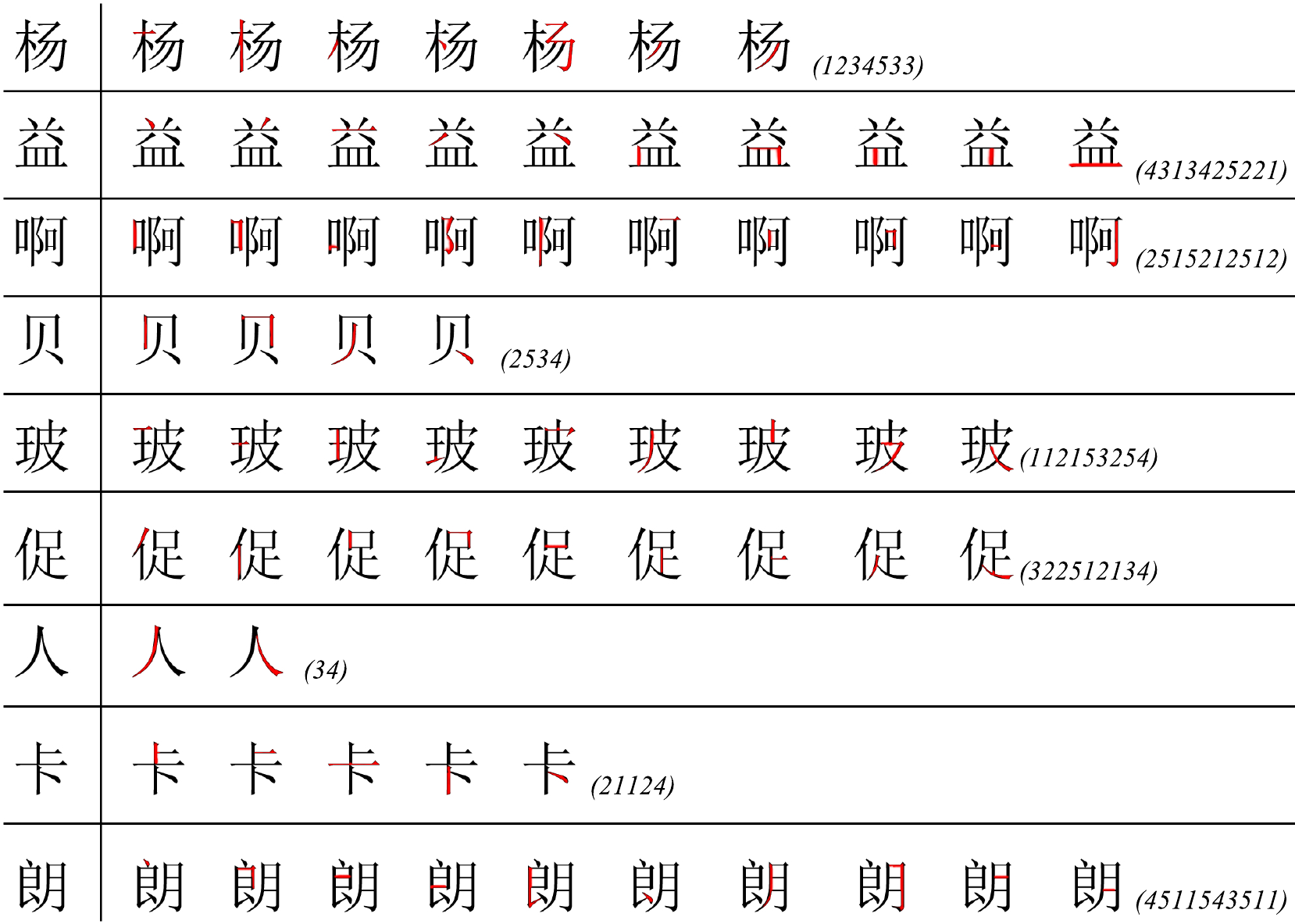}
    \caption{Nine examples of decomposition for Chinese characters.}
    \label{fig:nine examples}
\end{figure*}

To decompose Chinese characters, we follow the design in \cite{chen2021zero}. According to Unicode Han Database, there are five basic strokes in Chinese characters, including horizontal, vertical, left-falling, right-falling, turning. These five basic strokes are encoded from 1 to 5. Moreover, each Chinese character has its unique stroke sequence, which is available in Unicode Han Database and we display some examples in Figure \ref{fig:nine examples}. 


\begin{figure*}[t]
    \centering
    \includegraphics[width=1.0\textwidth]{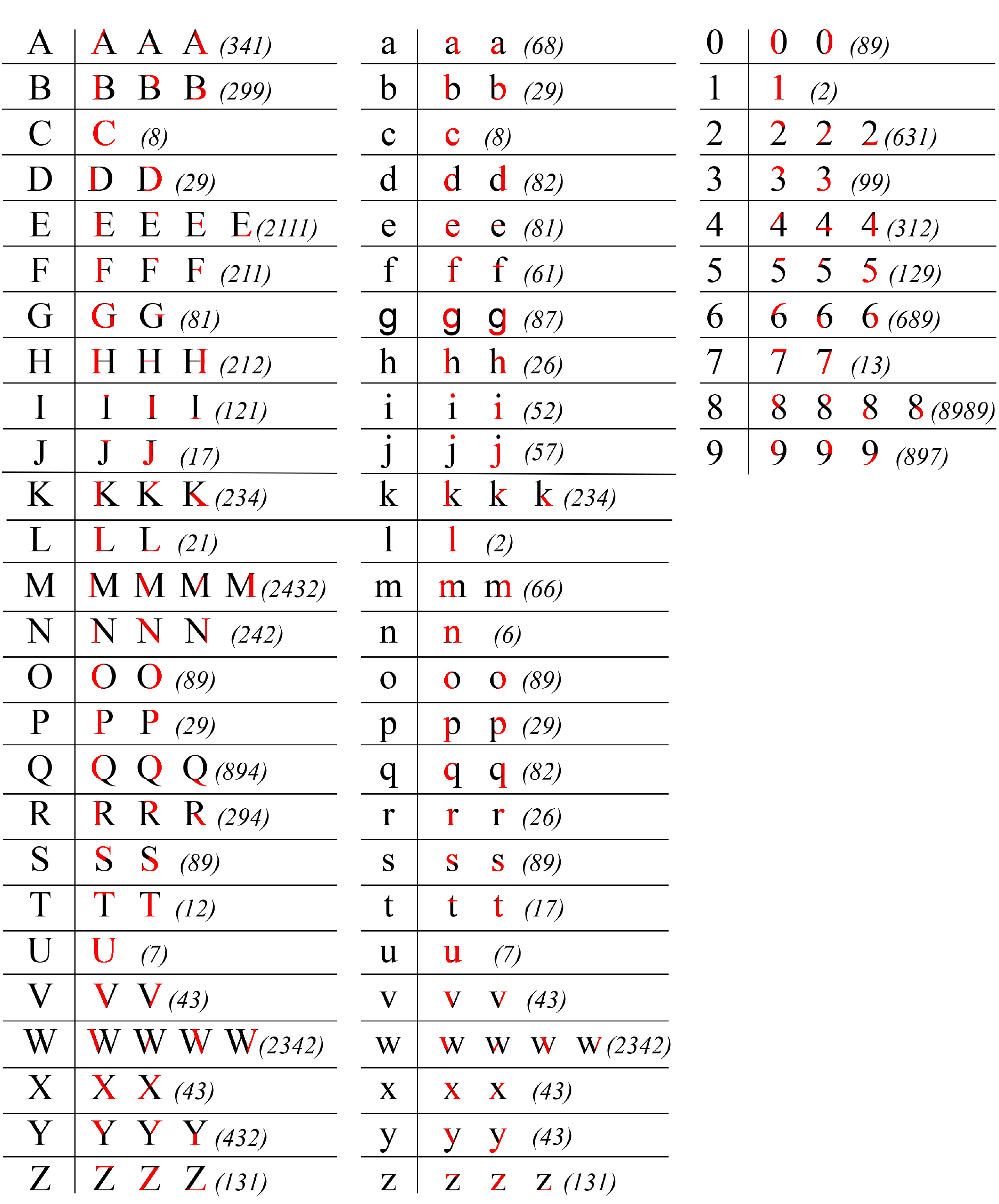}
    \caption{Decomposition of English characters and digits.}
    \label{fig:english_digit}
\end{figure*}

For English letters and digits, we manually design rules to decompose these characters, which are shown in Figure \ref{fig:english_digit}. Please note that the decomposition of several characters is not consistent with our writing habits since we decompose those ``turning'' strokes into tinier parts to pursue more fine-grained attention maps.

\section{Results of More Recognizers on TextZoom}
In addition to the experiments on recognizers of CRNN \cite{shi2016end}, MORAN \cite{luo2019moran} and ASTER \cite{shi2018aster}, we also conduct experiments on three more state-of-the-art recognizers (\eg NRTR \cite{sheng2019nrtr}, SEED \cite{qiao2020seed}, and AutoSTR \cite{zhang2020efficient}) to verify the ability of our method. The experimental results are shown in Table \ref{tab:final result}. We notice that the proposed SFM can indeed boost the recognition accuracy. For example, when using TSRN as the backbone and NRTR for evaluation, the super-resolution methods focusing at the stroke-level boost the average accuracy of 6.9\% compared with the one focusing at the character level. 

\begin{table*}[htb]
\caption{The experimental results of another three recognizers on TextZoom. According to the recognition accuracy, the module can indeed generate more recognizable text images with the guidance of SFM. The \textit{\underline{underlined}} numbers indicate the best average accuracy using the specific backbone and recognizer for evaluation. The \textbf{bold} numbers indicate the best average accuracy using the specific backbone and recognizer for evaluation.}
\centering
\scalebox{0.805}{
\begin{tabular}{c|c||c c c c|c c c c|c c c c}
\hline 
\multirow{2}*{Backbone} & \multirow{2}*{Focus} &  \multicolumn{4}{c|}{NRTR \cite{sheng2019nrtr}} & \multicolumn{4}{c|}{SEED \cite{qiao2020seed}} & \multicolumn{4}{c}{AutoSTR \cite{zhang2020efficient}}\\
\cline{3-14}
~ & ~ & Easy & Medium & Hard & Average & Easy & Medium & Hard & Average & Easy & Medium & Hard & Average  \\ 
\hline
\hline 
LR & -  & 65.1\% & 46.2\% & 31.8\% & 48.8\% & 70.2\% & 48.1\% & 34.8\% & 52.2\% & 67.8\% & 46.2\% & 32.9\% & 50.1\%  \\
\hline
HR & - & 91.9\% & 88.1\% & 76.5\% & 85.9\% & 94.0\% & 88.0\% & 77.5\% & 87.0\% & 93.6\% & 88.4\% & 78.5\% & 87.3\%  \\
\hline
\hline
\multirow{3}*{SRCNN}  & -  & 60.0\%  & 39.3\%  & 28.6\%  & 43.7\%  & 71.5\%  & 44.4\%  & 32.3\%  & 50.7\%  & 69.8\%  & 43.6\%  & 31.6\%  & 49.6\%   \\
~ & Char  & 59.1\% & 39.5\% & 28.6\% & 43.4\% & 70.3\% & 46.4\% & 33.4\% & 51.3\% & 67.9\% & 44.8\% & 31.9\% & 49.4\%  \\
~ & Stroke  & 64.9\% & 46.9\% & 31.9\% & \underline{49.0\%} & 71.3\% & 50.1\% & 36.5\% & \underline{53.8\%}  & 70.0\% & 49.9\% & 34.0\% & \underline{52.5\%}  \\
\hline
\multirow{3}*{SRResNet} & -  & 60.1\% & 48.8\% 
& 33.4\% & 48.3\% & 70.2\% & 56.8\% & 38.5\% & 56.1\% & 70.7\% & 55.1\% & 37.2\% & 55.4\%    \\
~ & Char  & 63.1\% & 49.6\% & 34.8\% & 50.1\% & 73.2\% & 58.0\% & 39.5\% & 57.9\% & 73.3\% & 57.2\% & 39.7\% & 57.8\%  \\
~ & Stroke  & 72.1\% & 56.5\% & 38.5\% & \underline{56.7\%} & 76.3\% & 58.0\% & 39.5\% & \underline{59.1\%}  & 76.6\% & 57.4\% & 38.9\% & \underline{58.8\%}  \\
\hline
\multirow{3}*{TBSRN} & -  & 65.3\%  & 49.0\%  & 37.2\%  & 51.4\%  & 74.6\%  & 56.2\%  & 41.8\%  & 58.6\%  & 74.6\%  & 55.1\%  & 41.5\%  & 58.1\%   \\
~ & Char  & 67.6\% & 52.2\% & 37.1\% & 53.3\% & 76.8\% & 58.3\% & 40.4\% & 59.7\% & 76.8\% & 59.5\% & 41.8\% & 60.5\%  \\
~ & Stroke  & \textbf{75.4\%} & 57.8\% & 41.3\% & \underline{59.2\%} & 76.5\% & 60.1\% & 42.4\% & \underline{60.7\%}  & 77.3\% & \textbf{61.2\%} & 41.8\% & \underline{61.2\%}  \\
\hline
\multirow{3}*{TSRN} & -  & 64.4\%  & 50.0\%  & 34.9\%  & 50.7\%  & 74.8\%  & 56.3\%  & 39.2\%  & 57.9\%  & 75.2\%  & 56.1\%  & 39.7\%  & 58.1\%   \\
~ & Char  & 66.8\% & 53.6\% & 37.3\% & 53.5\% & 75.8\% & 60.1\% & 41.0\% & 60.0\% & 75.7\% & 59.2\% & 40.2\% & 59.5\%  \\
~ & Stroke  & 75.3\% & \textbf{60.5\%} & \textbf{42.3\%} & \underline{\textbf{60.4\%}} & \textbf{78.6\%} & \textbf{61.3\%} & \textbf{43.1\%} & \underline{\textbf{62.1\%}}  & \textbf{77.4\%} & \textbf{61.2\%} & \textbf{42.3\%} & \underline{\textbf{61.4\%}}  \\
\hline
\end{tabular}}
\label{result on recognizer3}
\label{tab:final result}
\end{table*}

\section{Experimental Results of PSNR and SSIM}

\begin{table*}[t]
\caption{The experimental results of PSNR and SSIM.}
\centering
\scalebox{1}{
\begin{tabular}{p{2.5cm}<{\centering}|p{2.5cm}<{\centering}||p{2.5cm}<{\centering} p{2.5cm}<{\centering} p{2.5cm}<{\centering}}
\hline 
\multirow{2}*{Backbone} & \multirow{2}*{Foucs}  &  \multicolumn{3}{c}{PSNR/SSIM} \\
\cline{3-5}
~ & ~ & Easy & Medium & Hard   \\ 
\hline 
\hline
BICUBIC & - & 22.35/0.7884 & 18.98/0.6254 & 19.39/0.6592   \\
\hline
\multirow{3}*{SRCNN} & -  & 23.13/0.8152 & 19.57/0.6425 & 19.56/0.6833    \\
\cline{2-2}
~ & Char & 22.51/0.7952 & \textbf{19.66}/0.6288 & 19.44/0.6694\\
\cline{2-2}
~ & Stroke & 21.49/0.7532 & 19.56/0.6188 & 19.35/0.6404 \\
\hline

\multirow{3}*{SRResNet}  & -  & 20.65/0.8176 & 18.90/0.6324 & 19.53/0.7060  \\
\cline{2-2}
~ & Char & 23.16/0.8311 & 19.25/0.6314 & 19.94/0.6999 \\
\cline{2-2}
~ & Stroke & 23.16/0.8220 & 19.39/0.6316 & 20.04/0.6870 \\
\hline

\multirow{3}*{TSRN}  & -  & 22.95/0.8562 & 19.26/0.6596 & 19.76/0.7285  \\
\cline{2-2}
~ & Char & 23.34/0.8466 & 19.16/0.6379 & 19.81/0.7023 \\
\cline{2-2}
~ & Stroke & 23.34/0.8369 & 19.66/0.6499 & 19.90/0.6986  \\
\hline

\multirow{3}*{TBSRN}  & - & \textbf{24.13}/\textbf{0.8729} & 19.08/0.6455 & \textbf{20.09}/0.7452   \\
\cline{2-2}
~ & Char & 23.82/0.8660 & 19.17/0.6533 & 19.68/\textbf{0.7490}  \\
\cline{2-2}
~ & Stroke & 23.87/0.8674 & 19.43/\textbf{0.6644} & 19.95/0.7274   \\
\hline
\end{tabular}
}
\label{tab:psnr and ssim}
\end{table*}

\begin{figure*}[t]
    \centering
    \includegraphics[width=0.9\textwidth]{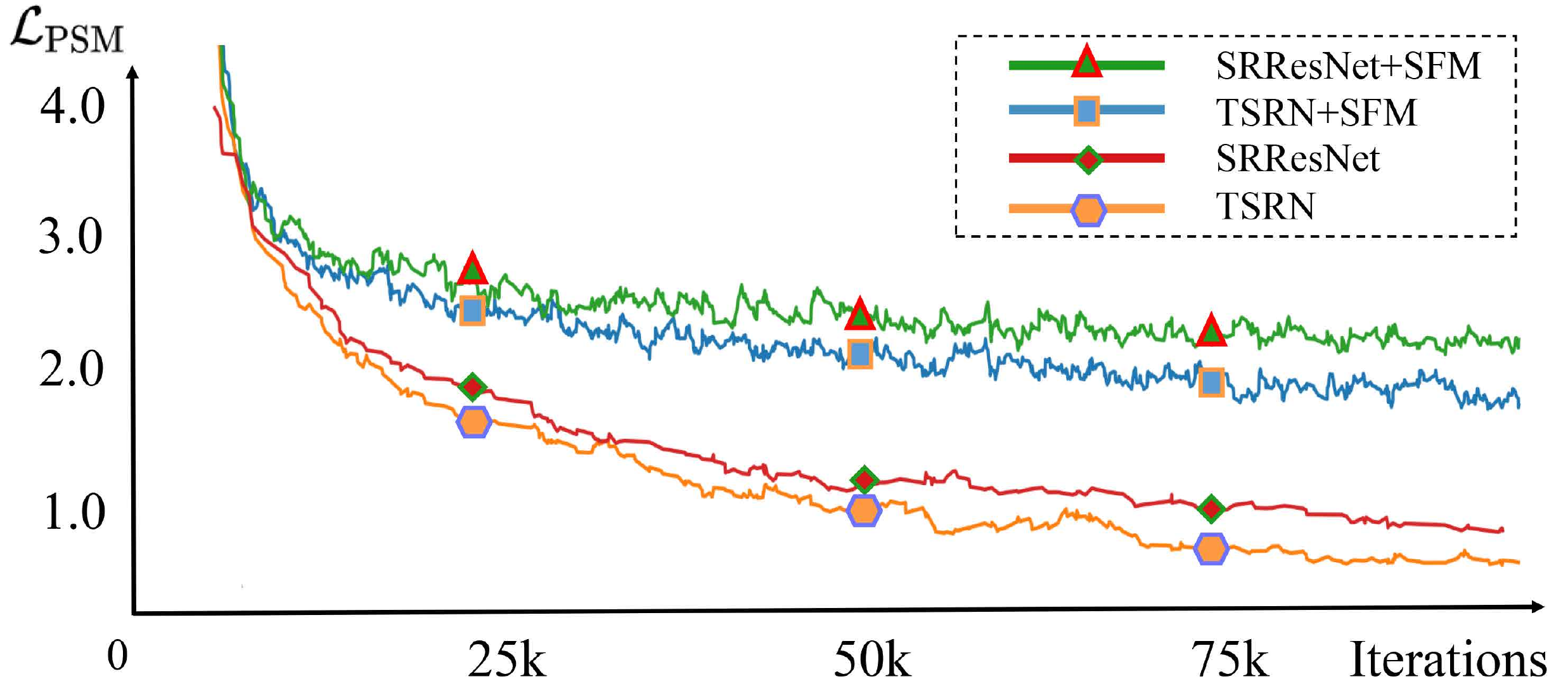}
    \caption{The trend of $L_{\text{PSM}}$ with different backbones.}
    \label{fig:loss_trend}
\end{figure*}

As is shown in Table \ref{tab:psnr and ssim}, when we employ PSNR and SSIM to evaluate the quality of generated images, we notice that the models with SFM perform poorly compared with their counterparts. We analyze the trend of $L_{\text{PSM}}$, which is shown in Figure \ref{fig:loss_trend}. According to this figure, we come to the following conclusions: the Position-wise Super-resolution Module usually pays attention to all areas of low-resolution images, whereas the proposed SFM mainly focuses on stroke-level details. Therefore, $L_{\text{PSM}}$ of models can drop quickly and show better convergence. In fact, the foreground (text regions) should be highlighted rather than the background in recovering the text images, especially in the text image super-resolution tasks. Moreover, as is mentioned in \cite{wang2020scene}, the image quality metrics are less important than the recognition accuracy. 

\section{Experimental Results on Public Benchmarks}
The dataset of IC15 \cite{karatzas2015icdar} includes several low-resolution images which may impose difficulty on the existing recognizers. We extract 352 low-resolution images from IC15 as a subset named IC15-352 and test on six recognizers, including CTC-based CRNN \cite{shi2016end}, rectification-based MORAN \cite{luo2019moran}, ASTER \cite{shi2018aster}, Transformer-based NRTR \cite{sheng2019nrtr}, semantics-based SEED \cite{qiao2020seed},  and NAS-based Auto-STR \cite{zhang2020efficient}. Although some semantics-based methods are capable of tackling those low-resolution images guided by exterior semantic knowledge, they still have difficulty in recognizing low-resolution images with global missing details. To verify the effectiveness of the proposed super-resolution method as a pre-processor, we conduct the experiments in three settings, including training TSRN without focus, with character-level focus, and with stroke-level focus. The experimental results are shown in Table \ref{tab:ic15}. Specifically, the model with stroke-level guidance boosts the accuracy of 3.1\% compared with that with character-level guidance. We also collect more low-resolution images from IC13 \cite{lucas2005icdar}, SVT \cite{wang2011end}, SVTP \cite{phan2013recognizing}, CUTE \cite{risnumawan2014robust}, resulting in 211 images, and we evaluate our method on the collected 211 images using PlugNet \cite{yanplugnet}, which is jointly trained with an SR branch. After super-resolution, the recognition accuracy can be boosted by 2.9\% (from 79.1\% to 82.0\%), which further validates the superiority of our method.

\begin{table*}[htb]
\caption{The experiment to validate the ability of our method as a pre-processor on IC15.}
\centering
\scalebox{0.975}{
\begin{tabular}{p{1.6cm}<{\centering}|p{1.6cm}<{\centering}||p{1.6cm}<{\centering}|p{1.6cm}<{\centering}|p{1.6cm}<{\centering}|p{1.6cm}<{\centering}|p{1.6cm}<{\centering}|p{1.6cm}<{\centering}}
\hline 
Method & Focus & CRNN & MORAN & ASTER & NRTR & AutoSTR & SEED  \\
\hline
Baseline & - & 43.8\% & 64.2\% & 68.5\%  & 65.6\% & 70.7\% & 71.9\% \\
\hline
\multirow{3}*{TSRN} & - & 54.0\% & 64.8\% & 71.0\% & 66.8\% & 70.2\% & 70.5\%\\
\cline{2-8}
~ & Char & 61.1\% & 71.6\% & 71.0\% & \textbf{70.2\%} & 72.4\% & 73.6\%\\
\cline{2-8}
~ & Stroke & \textbf{64.2\%} & \textbf{73.6\%} & \textbf{71.9\%} & 69.3\% & \textbf{73.6\%} & \textbf{74.4\%}\\
\hline
\end{tabular}
}
\label{tab:ic15}
\end{table*}

\begin{figure*}[h]
    \centering
    \includegraphics[width=0.9\textwidth]{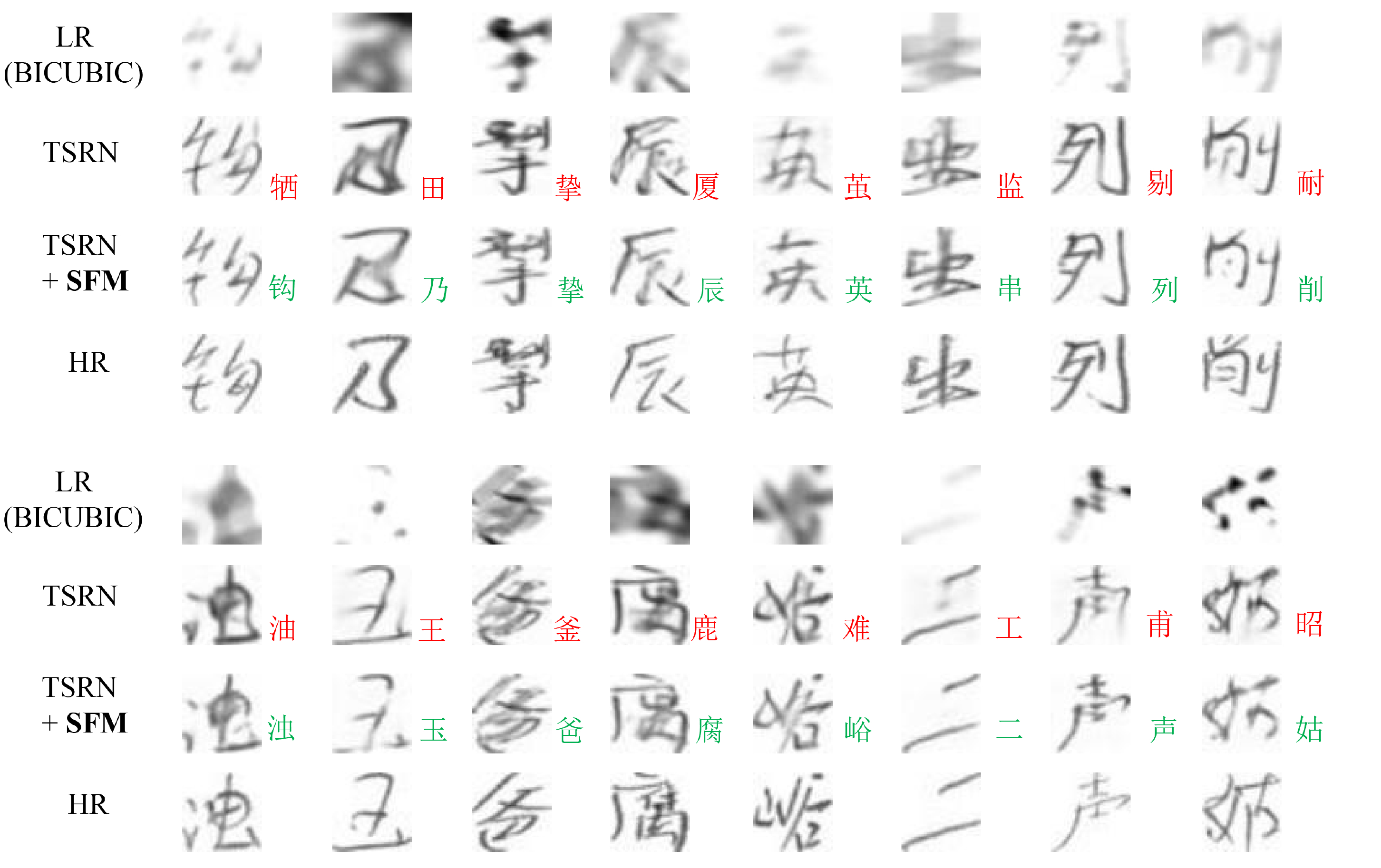}
    \caption{More Examples of Generated Images in ICDAR2013-HCCR.}
    \label{fig:ic13_more}
\end{figure*}

\end{document}